\documentclass{article}

\usepackage{arxiv}

\usepackage[utf8]{inputenc} 
\usepackage[T1]{fontenc}    
\usepackage{hyperref}       
\usepackage{url}            
\usepackage{booktabs}       
\usepackage{amsfonts}       
\usepackage{nicefrac}       
\usepackage{microtype}      
\usepackage{lipsum}
\usepackage{xcolor}
\usepackage{listings}
\usepackage{bm}
\usepackage{graphicx}
\usepackage[linesnumbered,vlined]{algorithm2e}

\definecolor{codegreen}{rgb}{0,0.5,0}
\definecolor{codepurple}{rgb}{0.88,0,0.32}
\definecolor{backcolour}{rgb}{0.97,0.97,0.97}
 
\lstdefinestyle{coding}{
    backgroundcolor=\color{backcolour},   
    commentstyle=\color{codegreen},
    keywordstyle=\color{codepurple},
    basicstyle=\ttfamily,
    breakatwhitespace=false,         
    breaklines=true,                 
    captionpos=b,                    
    keepspaces=true,                 
    numbers=none,                    
    numbersep=5pt,                  
    showspaces=false,                
    showstringspaces=false,
    showtabs=false,                  
    tabsize=4
}
 
\title{OPFython: A Python-Inspired Optimum-Path Forest Classifier}

\author{
  Gustavo H. de Rosa, João P. Papa \\
  Department of Computing \\
  São Paulo State University \\
  Bauru, São Paulo - Brazil \\
  \texttt{gustavo.rosa@unesp.br, joao.papa@unesp.br} \\
  \And
  Alexandre X. Falc\~ao \\
  Institute of Computing \\
  University of Campinas \\
  Campinas, S\~ao Paulo - Brazil \\
  \texttt{afalcao@ic.unicamp.br} \\
}
\begin{document}

\maketitle

\begin{abstract}
Machine learning techniques have been paramount throughout the last years, being applied in a wide range of tasks, such as classification, object recognition, person identification, and image segmentation. Nevertheless, conventional classification algorithms, e.g., Logistic Regression, Decision Trees, and Bayesian classifiers, might lack complexity and diversity, not suitable when dealing with real-world data. A recent graph-inspired classifier, known as the Optimum-Path Forest, has proven to be a state-of-the-art technique, comparable to Support Vector Machines and even surpassing it in some tasks. This paper proposes a Python-based Optimum-Path Forest framework, denoted as OPFython, where all of its functions and classes are based upon the original C language implementation. Additionally, as OPFython is a Python-based library, it provides a more friendly environment and a faster prototyping workspace than the C language.
\end{abstract}


\section{Introduction}
\label{s.intro}

Artificial Intelligence has become one of the most fostered research areas throughout the last years~\cite{Russell:16}. It is common to observe an increasing trend of automating tasks~\cite{Acemoglu:18}, which also fosters minimal human interaction algorithms, denoted as Machine Learning.

Machine learning research consists of developing new types of algorithms that do not need explicit instructions, relying on patterns and inferences~\cite{Bishop:06}. Also, they are designed in a way that humans can be assisted in decision-making tasks or even in daily activities automation, such as data retrieval~\cite{Manning:08}, intelligent gadgets~\cite{Li:15}, self-driving cars~\cite{Shalev:17}, among others. In the past decades, most machine learning-based algorithms were developed as symbolic- and knowledge-based models due to the difficulty in dealing with probabilistic models at that time~\cite{Shavlik:91}. Nevertheless, with the advent of computational power, probabilistic models were put in the spotlight as the availability of digitized information was no longer a problem~\cite{Langley:11}. Hence, most of today's algorithms rely on mathematical models and data sampling, i.e., models that are capable of learning occult patterns in training data and predicting unseen data, and are to a wide variety of tasks, such as computer vision~\cite{Sebe:05} and natural language processing~\cite{Indurkhya:10}.

\begin{figure*}
\centering
\begin{tabular}{cc}
    \includegraphics[scale=0.6]{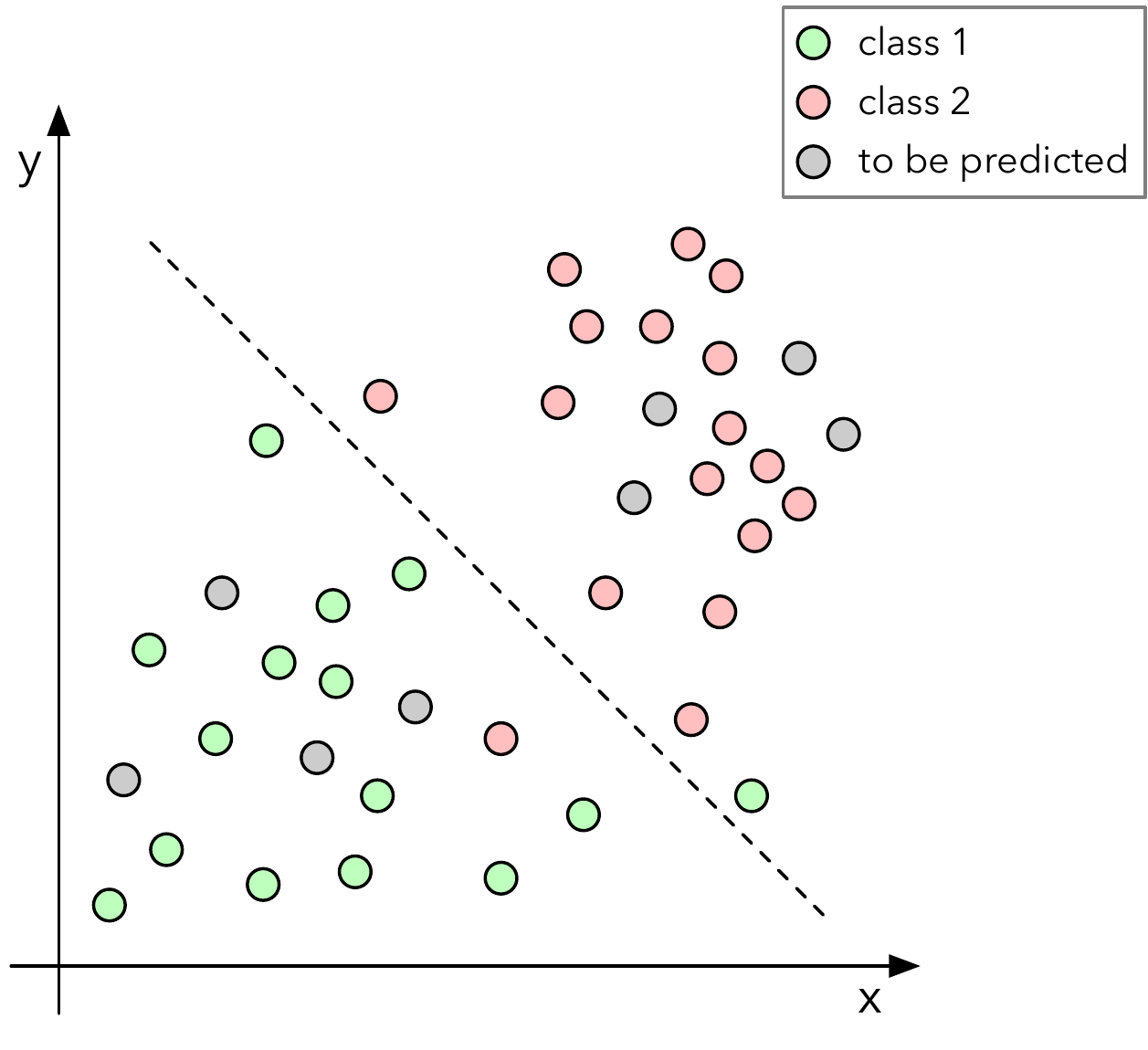} &
    \includegraphics[scale=0.6]{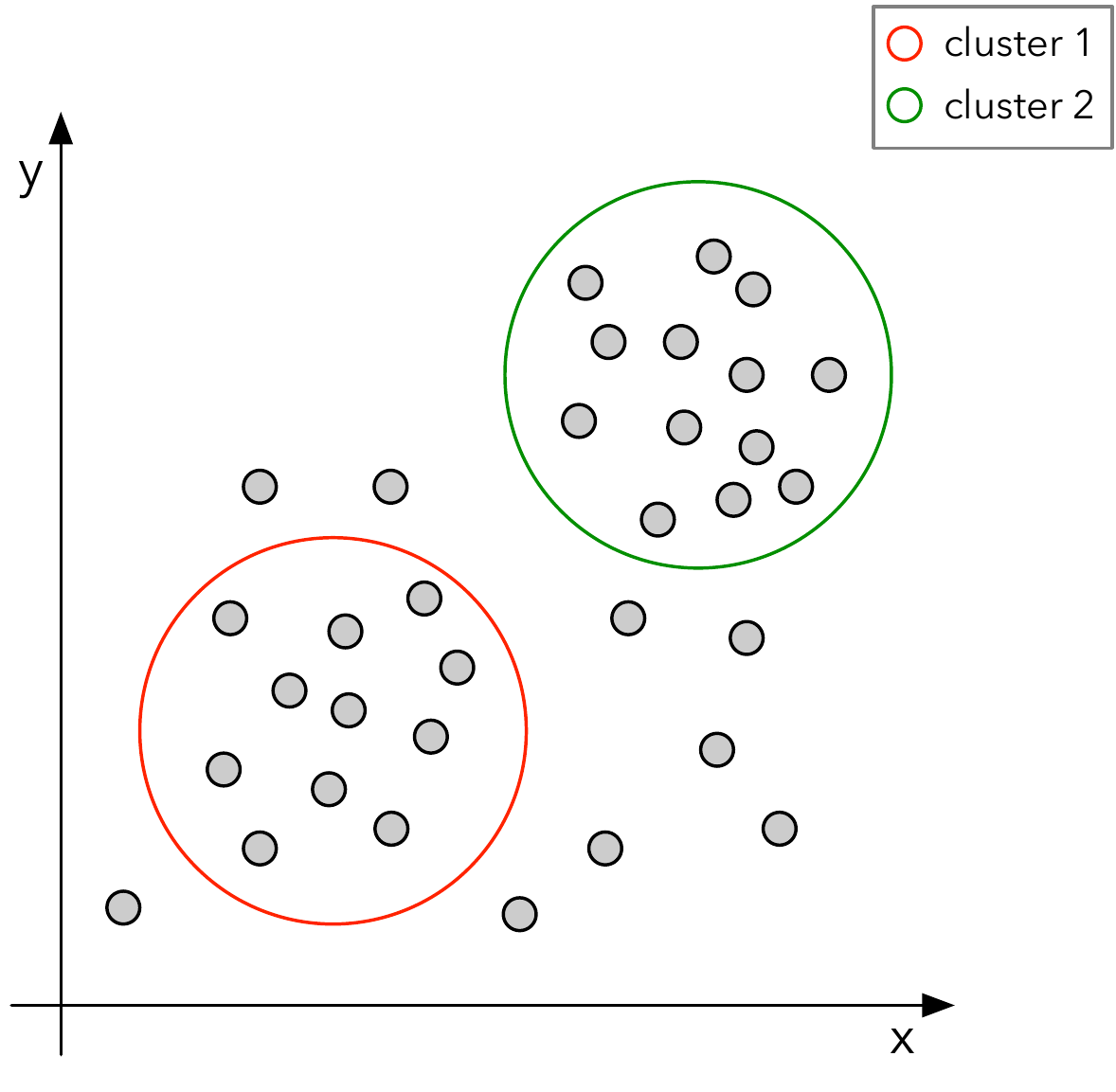} \\
    (a) & (b)
\end{tabular}
\label{f.tasks}
\caption{Illustration of the aforementioned machine learning types of learning: (a) supervised learning and (b) unsupervised learning.}
\end{figure*}

One can observe that it is possible to divide machine learning algorithms into two types of learning: supervised learning~\cite{Kotsiantis:07} and unsupervised learning~\cite{Hastie:09}, as depicted in Figure~\ref{f.tasks}. Concerning the supervised learning ones, such as classification and regression tasks, the algorithms aim to build mathematical models from labeled data, i.e., data containing the input features and the possible outputs (classes), and perform predictions on unseen data. Regarding unsupervised learning, the algorithms aim to build mathematical models capable of aggregating sets of data with common characteristics, known as clusters. In other words, unsupervised learning can discover patterns in data and group them into categories without knowing their actual labels.

Furthermore, it is crucial to observe that the machine learning area is closely related to other fields, such as data mining, optimization, and statistics. Regarding data mining, while machine learning focus on predicting information based on already-known properties from the data, the data mining area focuses on finding new properties in the data and transforming new knowledge into real-knowledge, for further application in machine learning algorithms~\cite{Friedman:98}. Concerning the optimization area, it is common to observe that most machine learning algorithms models are formulated as optimization problems, where some loss function is minimized over a set of training data. Essentially, the loss function is capable of expressing the discrepancy between the model's predictions and the actual samples, assisting the algorithm in learning the data's patterns and being capable of predicting unseen information~\cite{Sra:12}. Finally, regarding the statistics field, it is possible to perceive that statistics focus on drawing inferences from samples, while machine learning focuses on finding generalizable prediction patterns~\cite{Bzdok:18}. Additionally, due to their intimate relationship, some researchers combined machine learning and statistical methods in a new field of study, known as statistical learning~\cite{James:13}.

Recently, a new graph-based classifier proposed by Papa et al.~\cite{Papa:09}, known as Optimum-Path Forest (OPF), attempts to fulfill the literature with a parameterless classifier, which is effective during the learning step and efficiently when performing new predictions. Several works introduced the capacity of OPF and its state-of-the-art performance, being comparable to the well-known Support Vector Machines (SVM)~\cite{Chang:11} in supervised~\cite{Papa:12} and unsupervised learning~\cite{Rosa:14} tasks. Additionally, it provides tools, such as graph-cutting and K-Nearest Neighbors (KNN) graphs~\cite{PapaKNN:09}, to reduce the training set size with negligible effects on the accuracy of the classification. Nevertheless, the problem arises because there is only one official implementation based on the C language, making it difficult to be integrated with other well-known machine learning frameworks. Furthermore, there is a Python-based trend in the machine learning community.

This paper proposes an open-source Python Optimum-Path Forest classification library, called OPFython\footnote{https://github.com/gugarosa/opfython}. Mainly, the idea is to provide a user-friendly environment to work with Optimum-Path Forest classifiers by creating high-level methods and classes, removing the burden of programming from the user at a mathematical level.  The main contributions of this paper are threefold: (i) to introduce an Optimum-Path Forest classification library in the Python language, (ii) to provide an easy-to-go implementation and user-friendly framework, and (iii) to fill the lack of research regarding Optimum-Path Forest classifiers.

The remainder of this paper is organized as follows. Section~\ref{s.literature} presents a literature review and related works concerning Optimum-Path Forest classifiers frameworks. Section~\ref{s.theory} introduces a theoretical background concerning the supervised and unsupervised Optimum-Path Forest classifiers. Section~\ref{s.opfython} introduces thoughts of the OPFython library, such as its architecture, and an overview of the included packages. Section~\ref{s.library} provides more profound notions about the library, such as how to install, how to understand its documentation, some pre-included examples, and how to perform unitary tests. Furthermore, Section~\ref{s.applications} presents vital knowledge about the usage of the library, i.e., how to run pre-defined examples and model a new experiment. Finally, Section~\ref{s.conclusion} states conclusions and future works.
\section{Literature Review and Related Works}
\label{s.literature}

Optimum-Path Forest classifiers have arisen as a new approach to tackle supervised and unsupervised problems. They offer a parameterless graph-based implementation capable of executing an effective learning procedure while being extremely efficient when performing new predictions. It is possible to find its usage in a wide range of applications, such as feature selection~\cite{Rodrigues:14}, image segmentation~\cite{Miranda:09, Cappabianco:12}, signals classification~\cite{Nunes:14, Luz:13}. For instance, Iliev et al. applied an Optimum-Path Forest classification using glottal features for spoken emotion recognition, achieving state-of-the-art results comparable to the SVM classifier. Moreover, Ramos et al.~\cite{Ramos:11} applied an OPF-based classification for detecting non-technical energy losses, achieving outstanding results comparable to state-of-the-art artificial intelligence techniques. Furthermore,  Fernandes et al.~\cite{Fernandes:19} proposed a probabilistic-driven OPF classifier for detecting non-technical energy losses, improving the baselines obtained by the standard OPF.

Even though numerous works in the literature fosters the Optimum-Path Forests, there are some gaps in works regarding frameworks or open-sourced libraries. There is only an official implementation provided by Papa et al.~\cite{LibOPF:15}, denoted as LibOPF, which does not provide straightforward tools for users to design new experiments or integrate with other frameworks. Additionally, it lacks documentation and test suites, which help users understand the code and implement new methods and classes. Moreover, the library is implemented in C language, making it extremely difficult to integrate with other frameworks or packages, primarily because the machine learning community is turning their attention to the Python language.

Therefore, OPFython attempts to fill the gaps concerning Optimum-Path Forest frameworks. It is purely implemented in Python and provides comprehensive documentation, test suites, and several pre-loaded examples. Furthermore, every line of code is commented, there are continuous integration tests for every new push to its repository, a great readme that teaches how to get started with the library and full-time maintenance and support.
\section{Theoretical Foundation}
\label{s.theory}

Before diving into OPFython's library, we present a theoretical foundation about the Optimum-Path Forest. In the next subsections, we mathematically explain how the supervised and unsupervised classifiers work.

\subsection{Supervised Optimum-Path Forest}
\label{ss.supervised}

The Optimum-Path Forest is a multi-class classifier developed by Papa et al.~\cite{Papa:09}, being efficient in the training step and effective in the testing stage. Its foremost ability is to segment the feature space without requiring massive volumes of data. Essentially, the OPF classifier is a graph, having two possible adjacent relations: a complete graph or a $KNN$ graph. The difference between both methods is the adjacency relation, the methodology to estimate the prototypes\footnote{Prototypes are master nodes representing a specific class and conquer other nodes.}, and the path cost function.

The principal idea behind the supervised OPF is to construct a complete graph, where any two samples are connected. In this case, the nodes represent the samples' features vector, and the edges connect all nodes. Regarding the prototypes, the same are chosen throughout Minimum Spanning Trees (MST)\footnote{MSTs are subgraphs that connect all nodes within the same set using the minimum possible cost.} To find the nearest samples from different classes, namely, by selecting samples located in the classes frontiers\footnote{Regions more likely to classification mistakes.}. After the prototypes definition, they compete to conquer adjacent nodes while trying to find the best path (lowest cost) defined by the path cost function and create Optimum-Path Trees (OPT). Finally, during the testing phase, OPF inserts each new sample into the graph and finds the prototype, which offers the minimum cost path (class labeling).

Let ${\mathcal Z}$ be a dataset, where ${\mathcal Z} = {\mathcal Z}_1\bigcup {\mathcal Z}_2$, and ${\mathcal Z}_1$ and ${\mathcal Z}_2$ represents the training and testing sets, respectively. Each sample $s \in {\mathcal Z}$ can be represented by its feature vector $\overrightarrow{v}(s) \in \Re^{n}$. OPF$_{cpl}$ graph is represented by ${\mathcal G} = ({\mathcal V},{\mathcal A})$, where ${\mathcal A}$ refers to the set of edges that connects all nodes pairs and ${\mathcal V}$ is the features vectors set $\overrightarrow{v}(s)$, $\forall s \in {\mathcal Z}$. In addition, let $\lambda (\cdot)$ be a function that assigns a real label for each sample in ${\mathcal Z}$.

\subsubsection{Training Step}
\label{sss.train} 

Let the graph ${\mathcal G}_1 = ({\mathcal V}_1, {\mathcal A})$ be inducted from the training set, where ${\mathcal V}_1$ holds all feature vectors from samples belonging to the training set. The first objective of the training phase is to obtain a set of prototypes ${\mathcal S}$, onde ${\mathcal S} \subset {\mathcal Z}_1$.

Let a path $\pi_s$, with ending in $\textbf{s}$, be in ${\mathcal G}$ and a function $f(\pi_s)$ that associates a value to this path. In order to a prototype conquer adjacent samples, the purpose is to minimize $f(\pi_s)$ through a path cost function given by the following equation:

\begin{eqnarray}
\label{eq:opf.fmax}
f_{max}(\langle
\textbf{s}\rangle) & = & \left\{ \begin{array}{ll}
  0 & \mbox{se $\textbf{s}\in S$,} \\
+\infty & \mbox{otherwise}
\end{array}\right. \nonumber \\
f_{max}(\pi \cdot \langle \textbf{s},\textbf{t} \rangle) & = & \max\{f_{max}(\pi),d(\textbf{s},\textbf{t})\}.
\end{eqnarray}

\noindent where $f_{max}(\pi \cdot \langle \textbf{s},\textbf{t} \rangle)$ computes the maximum distance between adjacent samples $\textbf{s}$ and $\textbf{t}$ along the path $\pi \cdot \langle \textbf{s},\textbf{t} \rangle$. A path $\pi_s$ is referred as optimum if $f(\pi_s) \leq f(\tau_s)$ for any other path $\tau_s$.

The minimization of $f_{max}$ assigns to each sample $\textbf{t} \in {\mathcal Z}_1$ an optimum path $P^\ast(\textbf{t})$, whose minimum cost $C(\textbf{t})$ is given by the following equation:

\begin{eqnarray}
\label{eq:opf.cost}
C(\textbf{t}) = \min_{\forall \pi_t \in ({\mathcal Z}_1, {\mathcal A})}\{f_{max}(\pi_t)\}.
\end{eqnarray}

\subsubsection{Testing Step}
\label{sss.test}

The testing set graph ${\mathcal G}_2 = ({\mathcal V}_2, {\mathcal A})$ is composed by samples $\textbf{t} \in {\mathcal V}_2$. Each sample \textbf{t} is connected to a sample $\textbf{s} \in {\mathcal V}_1$, making \textbf{t} as part of the original graph. The objective is to find an optimum path $P^\ast(\textbf{t})$ from ${\mathcal S}$ to \textbf{t} with class $\lambda(R(\textbf{t}))$ of its prototype $R(\textbf{t}) \in {\mathcal S}$. Consequently, the sample \textbf{t} is removed from the graph. This path can be identified by evaluating the optimum cost value $C(\textbf{t})$:

\begin{equation}
\label{eq:opf.opt_cost}
C(\textbf{t}) = \min\{{\max \{C(\textbf{s}),d(\textbf{s},\textbf{t})\}}\}, \forall \textbf{s} \in {\mathcal Z}_1.
\end{equation}

\subsection{Unsupervised Optimum-Path Forest}
\label{ss.unsupervised}

Let ${\cal N}$ be a dataset such that for every sample $s\in {\cal N}$ there is a feature vector $\bm{v}(s)$. Additionally, let $d(s,t)$ be the distance between samples $s$ and $t$ in the feature space, which is described by $d(s,t)=\|\bm{v}(t)-\bm{v}(s)\|$.

A graph ${\cal (N,A)}$ is defined by arcs $(s,t)\in {\cal A}$ that connect $k$-nearest neighbors in the feature space. The arcs are weighted by $d(s,t)$ and the nodes $s\in {\cal N}$ are weighted by a density value $\rho(s)$, given by Equation~\ref{e.density}:

\begin{eqnarray}
 \label{e.density}
  \rho(s) & = & \frac{1}{\sqrt{2\pi\sigma^2}|{\cal A}(s)|} \sum_{\forall t\in {\cal A}(s)} \exp\left(\frac{-d^2(s,t)}{2\sigma^2}\right),
\end{eqnarray}
where $|{\cal A}(s)|=k$, $\sigma = \frac{d_f}{3}$, and $d_f$ is the maximum arc weight in ${\cal (N,A)}$. Using these parameters, all nodes are considered for density computation, since a Gaussian function covers most samples within $d(s,t)\in [0,3\sigma]$.

A standard method to compute a probability density function (p.d.f.) is the Parzen-window. Equation (\ref{e.density}) provides a Parzen-window estimation based on an isotropic Gaussian kernel when the arcs $(s,t) \in {\cal A}$ if $d(s,t)\leq d_f$ are defined. Nevertheless, this approximation causes differences in scale and sample concentration.

An interesting way to solve this problem is to choose $d_f$ based on a particular region of the feature space~\cite{Comaniciu:03}. By considering the $k$-nearest neighbors, it is possible to handle different concentrations and transform the scale problem into finding the best $k$ value within $[1,k_{\max}]$, for $1 \leq k_{\max}\leq |{\cal N}|$. An approach proposed by Rocha et al.~\cite{RochaIJIST:09} considers the minimum graph cut (best $k$) according to a measure suggested by Shi et al.~\cite{Shi:00}.

Moreover, let a path $\pi_t$ be the sequence of adjacent samples starting from a root $R(t)$ and ending at a sample $t$, being $\pi_t=\langle t\rangle$ a trivial path and $\pi_s\cdot \langle s,t\rangle$ the concatenation of $\pi_s$ and arc $(s,t)$. Among all possible paths $\pi_t$ with roots on the maxima of the p.d.f., the problem lies in finding a path with the lowest density value. Each path defines an influence zone (cluster) by selecting strongly connected samples. Mathematically speaking, Equation~\ref{e.pf2} maximizes $f(\pi_t)$ for all $t\in {\cal N}$ where:

\begin{eqnarray}
f(\langle t \rangle) & = & \left\{ \begin{array}{ll} 
    \rho(t)           & \mbox{if $t \in {\cal R}$} \\
    \rho(t) - \delta  & \mbox{otherwise}
 \end{array}\right. \nonumber
\end{eqnarray}
and
\begin{eqnarray}
\label{e.pf2}
f(\langle \pi_s\cdot \langle s,t\rangle\rangle)&=& \min \{f(\pi_s), \rho(t)\},
\end{eqnarray}
where $\delta = \min_{\forall (s,t)\in {\cal A} | \rho(t) \neq \rho(s) }
|\rho(t)-\rho(s)|$ and ${\cal R}$ is a root with one element set for each maximum of the p.d.f.. One can see that higher values of $\delta$ reduces the number of maxima. Additionally, in this library, we are using $\delta=1.0$ and $\rho(t)\in [1,1000]$.

Finally, the OPF algorithm maximizes $f(\pi_t)$ such that the optimum paths compose an Optimum-Path Forest, i.e.,  a predecessor no-cycling map $P$ which assigns to each sample $t\notin {\cal R}$ its predecessor $P(t)$ from the optimum path ${\cal R}$ or a marker $nil$ when $t\in {\cal R}$. Each p.d.f. maximum (prototype) is the root of an OPT, commonly known as a cluster. Furthermore, the collection of all OPTs is the so-called Optimum-Path Forest.
\section{OPFython}
\label{s.opfython}

\begin{figure}[!ht]
\centering
\includegraphics[scale=0.55]{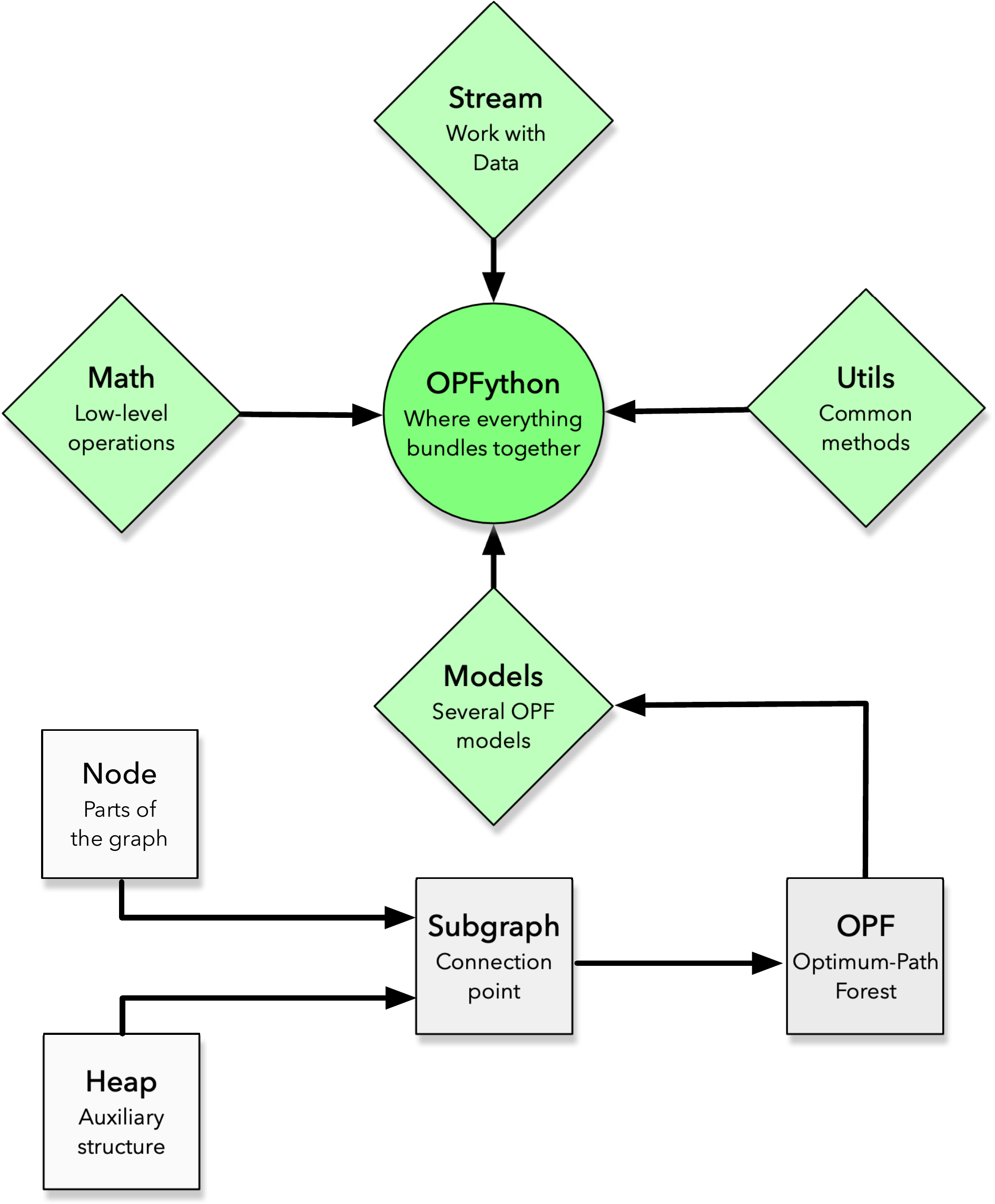}
\caption{Flowchart of OPFython's architecture.}
\label{f.flowchart}    
\end{figure}

OPFython is distributed among several packages, each one being accountable for particular classes and methods. Figure~\ref{f.flowchart} represents a summary of OPFython's architecture, while the next sections present each of its packages within more details.

\subsection{Core}
\label{ss.core}

\begin{figure}[!ht]
\centering
\includegraphics[scale=0.5]{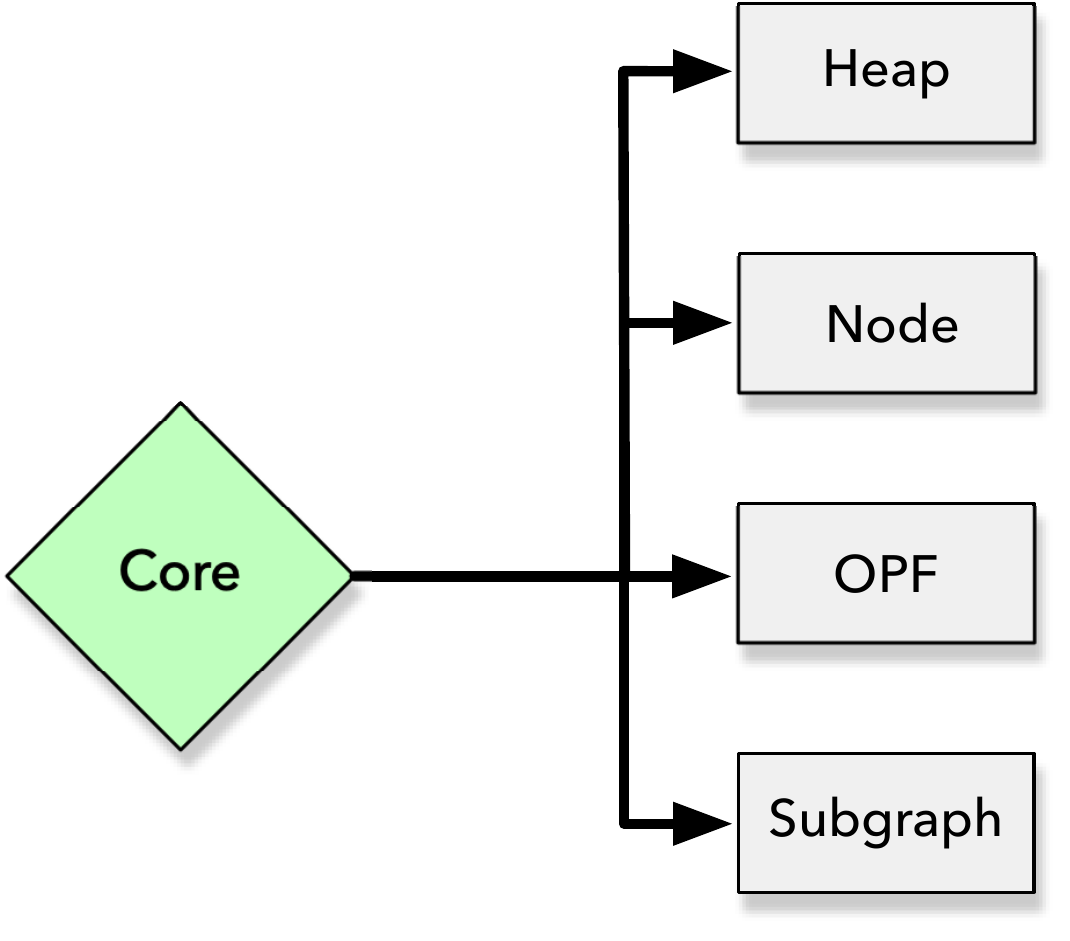}
\caption{Flowchart of OPFython's core package.}
\label{f.flowchart_core}    
\end{figure}

The core package serves as the origin of all OPFython's sub-classes. It assists as a building base for implementing more appropriate structures that one may require when creating an Optimum-Path Forest-based classifier. As portrayed in Figure~\ref{f.flowchart_core}, four modules compose the core package, as follows:

\begin{itemize}

\item \textbf{Heap:} The heap assists OPF in stacking nodes' according to their costs and further unstacking them to build the subgraph;

\item \textbf{Node:} When working with graph-based structures, each of their pieces is represented by a node. In OPFython, we use the node structure to store valuable information of a sample, such as its features, label, and other information that OPF might need;

\item \textbf{OPF:} The OPF class serves as the classifier itself. It implements some basic methods that are common to its children, as well as some methods that assist users in saving and loading pre-trained models;

\item \textbf{Subgraph:} The subgraph is one of the most fundamental structures of the OPF classifier. A graph-based classifier uses nodes and arcs to build the optimum-path costs and find the prototype nodes, which conquer the remaining samples and propagate their labels.

\end{itemize}

\subsection{Math}
\label{ss.math}

\begin{figure}[!ht]
\centering
\includegraphics[scale=0.5]{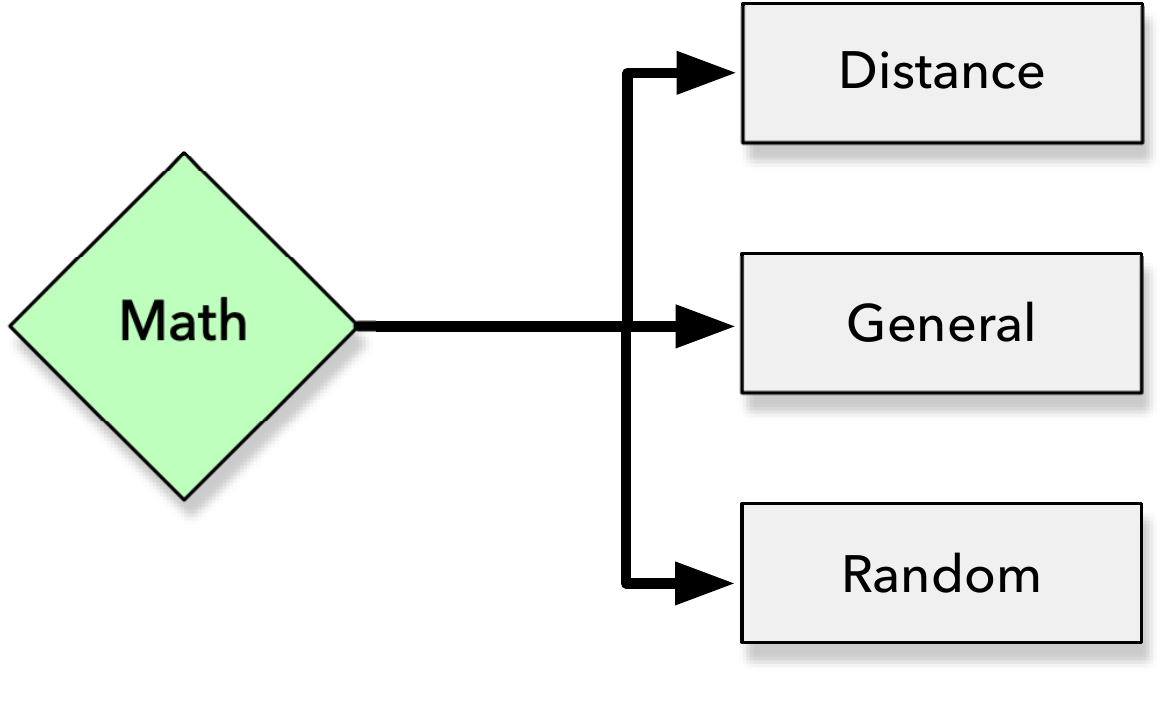}
\caption{Flowchart of OPFython's math package.}
\label{f.flowchart_math}    
\end{figure}

To ease the user's life, OPFython offers a mathematical package, containing low-level math implementations, illustrated by Figure~~\ref{f.flowchart_math}. Naturally, some repeated functions that are used throughout the library are represented in this package, as follows:

\begin{itemize}

\item \textbf{Distance:} A distance metric is used to calculate the cost between nodes. Hence, we offer a variety of distance metrics that fulfills every task needs;

\item \textbf{General:} Common-use functions that do not have a special division are defined in this module;

\item \textbf{Random:} Lastly, some methods might use random numbers for sampling or setting a heuristic. This module can generate uniform and Gaussian random numbers.

\end{itemize}

\subsection{Models}
\label{ss.models}

\begin{figure}[!ht]
\centering
\includegraphics[scale=0.5]{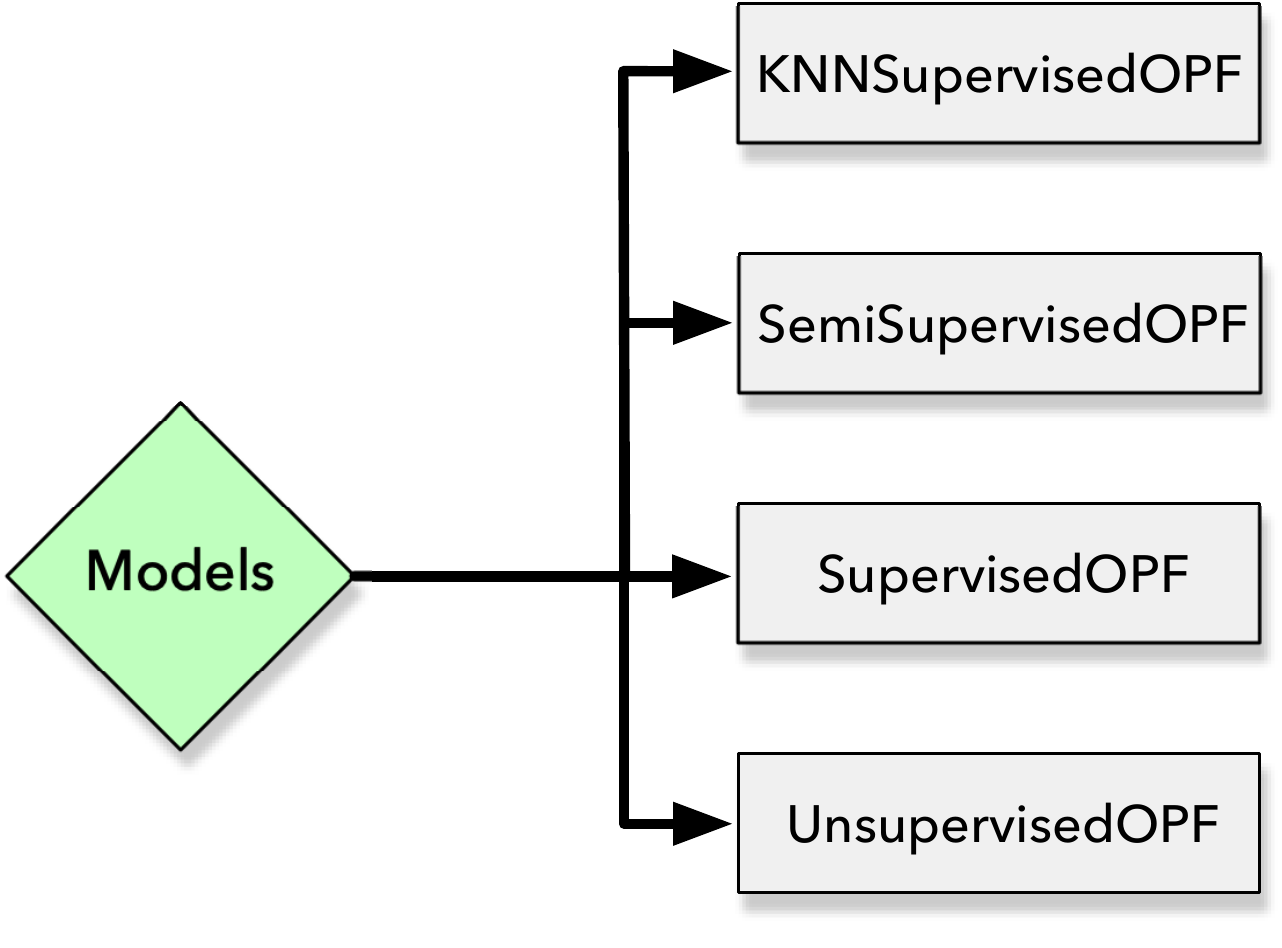}
\caption{Flowchart of OPFython's models package.}
\label{f.flowchart_models}    
\end{figure}

Several approaches are to be conducted when designing an optimum-path forest classifier, such as supervised, unsupervised, semi-supervised, among others. Therefore, the models' package provides classes and methods that compose these high-level abstractions and implement the classifying strategies. Currently, OPFython offers four types of classifiers, which are illustrated by Figure~\ref{f.flowchart_models} and described as follows:

\begin{itemize}

\item \textbf{KNNSupervisedOPF~\cite{PapaKNN:09}:} A supervised Optimum-Path Forest classifier that uses a KNN-based subgraph, providing a more effective way to build up the connectivity subgraph;

\item \textbf{SemiSupervisedOPF~\cite{Amorim:14}:} A semi-supervised Optimum-Path Forest classifier, which is extremely useful in labeling unknown samples;

\item \textbf{SupervisedOPF~\cite{Papa:09}:} The classical supervised Optimum-Path Forest classifier, which is suitable for training on labeled datasets and performing new predictions;

\item \textbf{UnsupervisedOPF~\cite{Rocha:09}:} The standard unsupervised Optimum-Path Forest classifier, which is suitable for clustering unlabeled datasets.

\end{itemize}

\subsection{Stream}
\label{ss.stream}

\begin{figure}[!ht]
\centering
\includegraphics[scale=0.5]{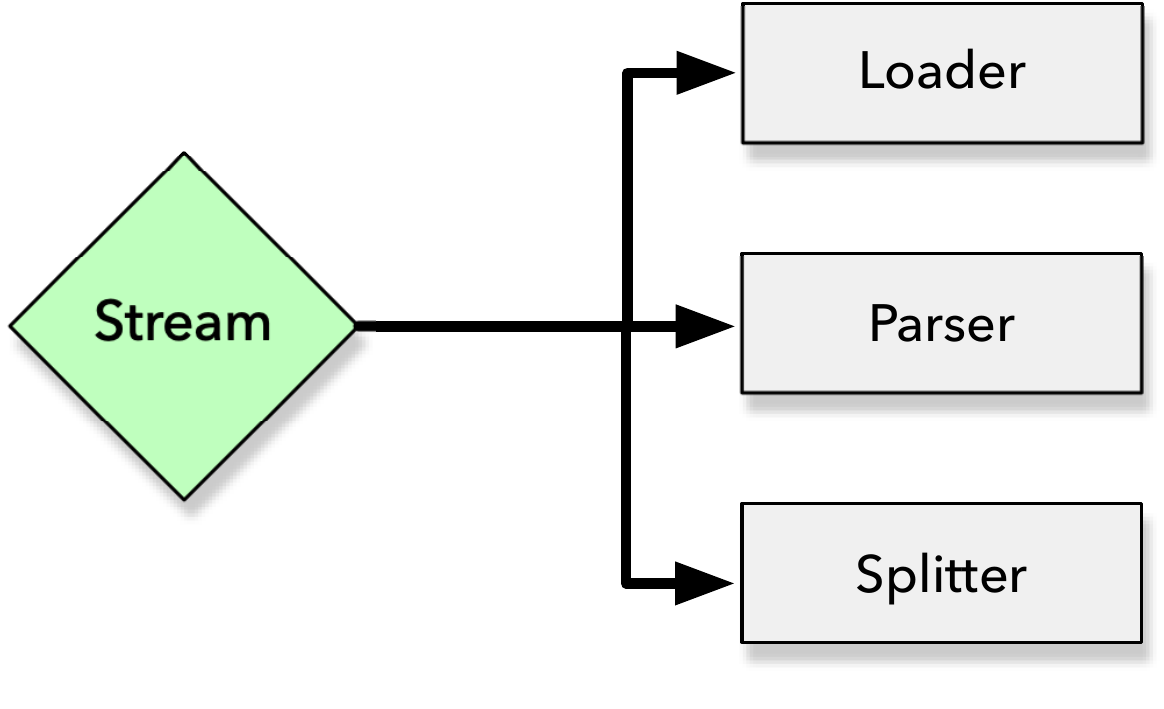}
\caption{Flowchart of OPFython's stream package.}
\label{f.flowchart_stream}    
\end{figure}

The stream package deals with every pre-processing step of the input data. It is essentially responsible for loading the data, parsing it into samples and labels, and splitting it into new sets, such as training, validation, and testing. Figure~\ref{f.flowchart_stream} depicts its modules, as well as we provide a brief description of them as follows:

\begin{itemize}

\item \textbf{Loader:} A loading module that assists users in pre-loading datasets. Currently, it is possible to load files in .txt, .csv and .json formats;

\item \textbf{Parser:} After loading the files, it is necessary to parse the pre-loaded arrays into samples and labels;

\item \textbf{Splitter:} Finally, if necessary, one can split the loaded and parsed dataset into new sets, such as training, validation, and testing.

\end{itemize}

\subsection{Subgraphs}
\label{ss.subgraphs}

\begin{figure}[!ht]
\centering
\includegraphics[scale=0.55]{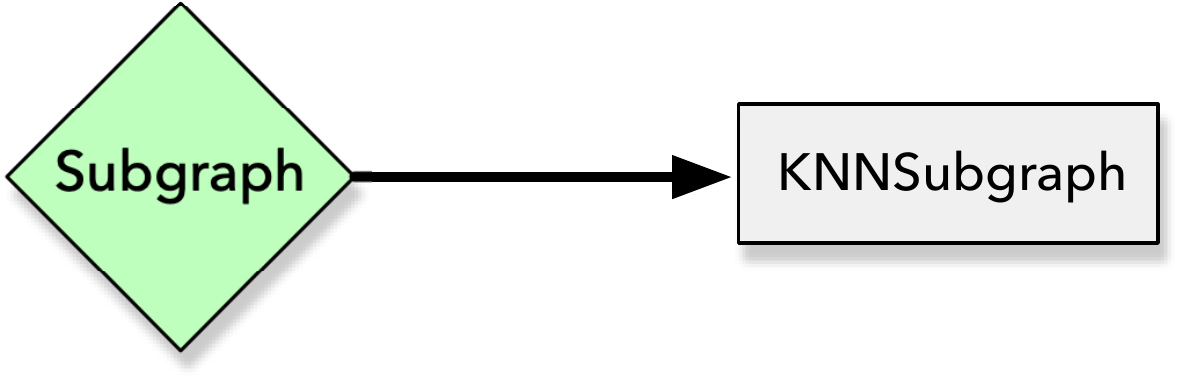}
\caption{Flowchart of OPFython's subgraphs package.}
\label{f.flowchart_subgraphs}    
\end{figure}

As mentioned before, the subgraph is one of the essential structures of the classification process. Nevertheless, one can observe that distinct classifiers might need distinct subgraphs. Therefore, we are glad to offer additional subgraphs implementations as portrayed by Figure~\ref{f.flowchart_subgraphs} and described as follows:

\begin{itemize}

\item \textbf{KNNSubgraph:} When dealing with KNN-based classifiers, it is crucial to use a KNN-based subgraph, as it implements some additional functions that the classifier might need.

\end{itemize}

\subsection{Utils}
\label{ss.utils}

\begin{figure}[!ht]
\centering
\includegraphics[scale=0.55]{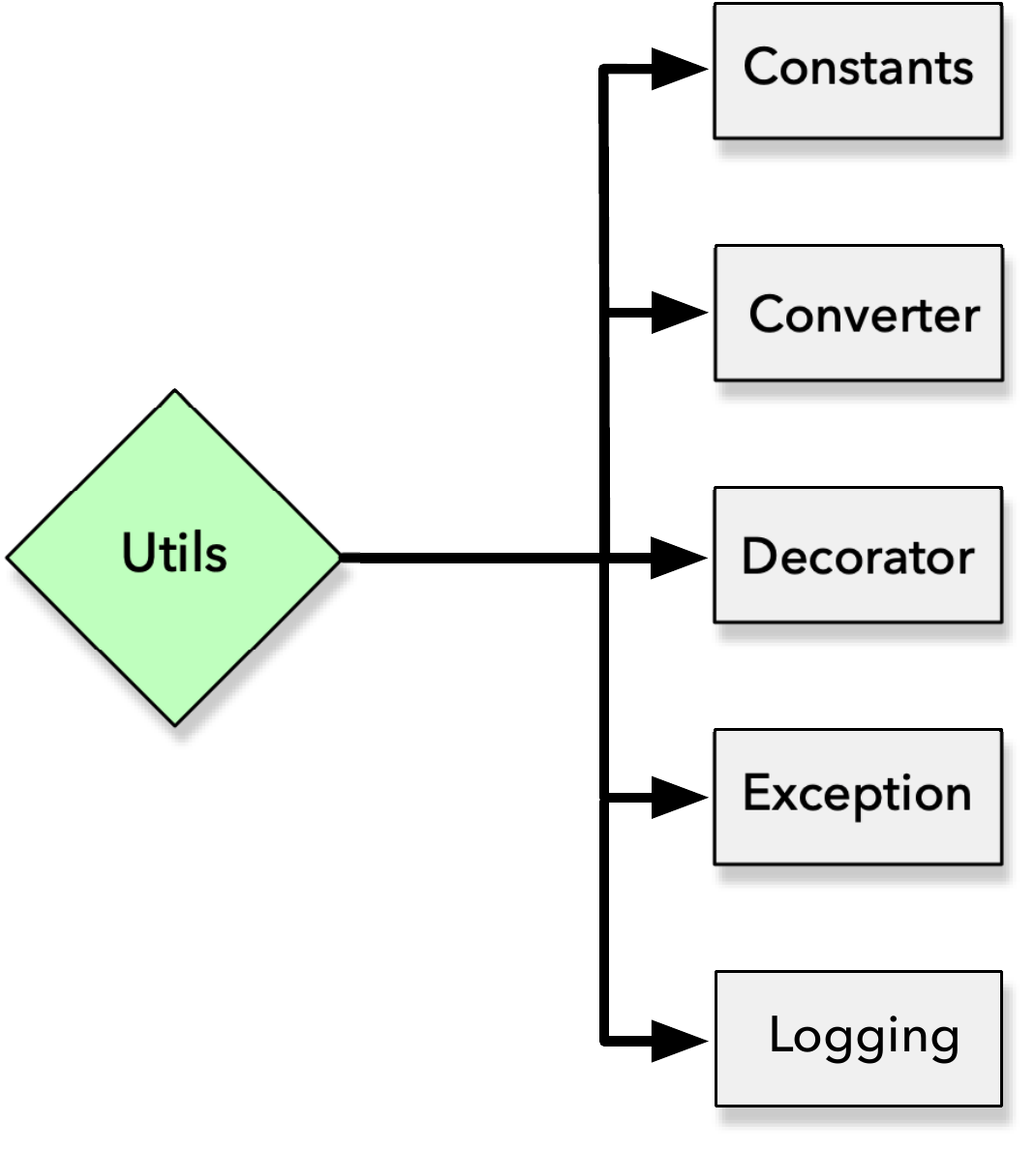}
\caption{Flowchart of OPFython's utils package.}
\label{f.flowchart_utils}    
\end{figure}

A utility package implements standard tools shared over the library, as it is a better approach to implement once and re-use them across other modules, as shown in Figure~\ref{f.flowchart_utils}. This package implements the subsequent modules:

\begin{itemize}

\item \textbf{Constants:} Constants are fixed numbers that do not alter throughout the code. For the sake of easiness, they are implemented in the same module;

\item \textbf{Converter:} Most of OPF users are familiarized to the specific file format it uses. Hence, we implement an own module that is capable of converting .opf files into .txt, .csv, and .json;

\item \textbf{Decorator:} Wrappers that provide common functionalities before running pieces of code;

\item \textbf{Exception:} In order to assist users, the exception module implements common errors and exceptions that might happen when invalid arguments are used in OPFython classes and methods;

\item \textbf{Logging:} Every method that is invoked in the library is logged onto a log file. One can watch the log to detect potential errors, essential warnings, or even success messages throughout the classification procedure.

\end{itemize}

\section{Library Usage}
\label{s.library}

In this section, we describe how to install the OPFython library and the first steps to start playing with it. Essentially, one can study its documentation or make use of the already-included examples. Besides, there are implemented methods that conduct unitary tests and verify if everything is operating as presumed.

\subsection{Installation}
\label{ss.installation}

First of all, we understand that everything has to be smooth without being tricky or daunting. Therefore, OPFython will always be the one-to-go package, from the very first installation to its further usage. Just execute the following command under the most preferred Python environment (standard, conda, virtualenv):

\verb|pip install opfython|

Alternatively, it is possible to use the bleeding-edge version by cloning its repository and installing it:

\verb|git clone https://github.com/gugarosa/opfython.git|

\verb|pip install .|

Note that there is no other requirement to use OPFython. As its single dependency is the Numpy package, it can be installed everywhere, despite the machine's operational system.

\subsection{Documentation}
\label{ss.documentation}

One might have an enthusiasm for mastering the concepts and strategies behind OPFython. Hence, we provide a fully documented reference\footnote{https://opfython.readthedocs.io} containing everything that the library offers. From elementary classes to more complex methods, OPFython's documentation is the perfect reference for learning how the library was developed or even improving it with contributions.

\subsection{Classes and Methods Examples}
\label{ss.example}

Additionally, in the \verb|examples/| folder, we provide example scripts for all packages that the library implements, such as:

\begin{itemize}

\item \textbf{Core:} \verb|create_heap.py|, \verb|create_node.py|, \verb|create_opf.py|, \verb|create_subgraph.py|;
\item \textbf{Math:} \verb|calculate_distances.py|, \verb|general_purpose.py|, \verb|generate_random_numbers.py|, \verb|pre_compute_distances.py|;
\item \textbf{Models:} \verb|create_knn_supervised_opf.py|, \verb|create_semi_supervised_opf.py|,\\ \verb|create_supervised_opf.py|, \verb|create_unsupervised_opf.py|;
\item \textbf{Stream:} \verb|load_file.py|, \verb|parse_loaded_file.py|, \verb|split_data.py|;
\item \textbf{Subgraphs:} \verb|create_knn_subgraph.py|.
\item \textbf{Utils:} \verb|convert_from_opf.py|;

\end{itemize}

Each example is constituted of high-level explanations of how to use predefined classes and methods. One can observe that it provides a standard description of how to instantiate each class and decide which arguments should be employed.

\subsection{Test Suites}
\label{ss.test}

\begin{figure}[!ht]
\centering
\includegraphics[scale=1]{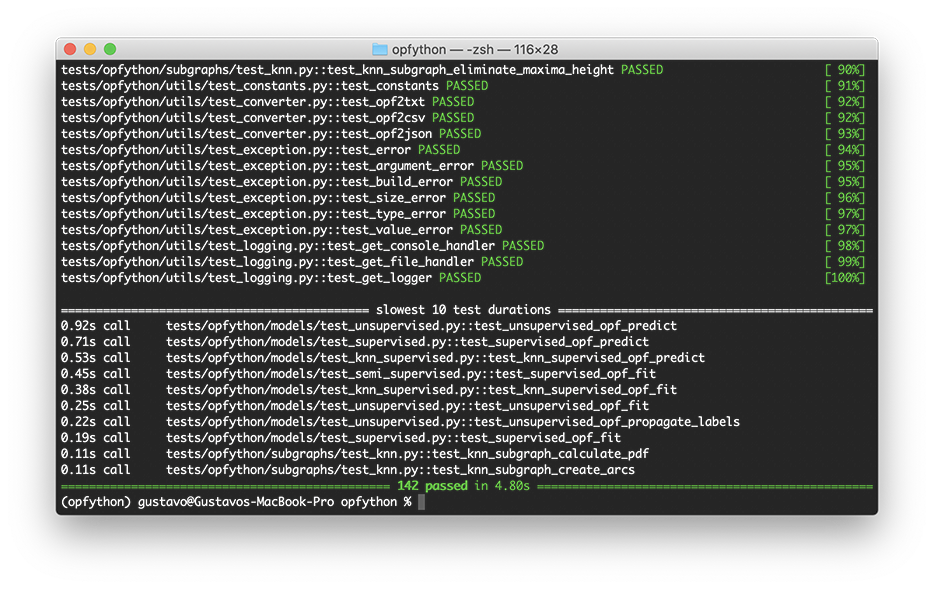}
\caption{Example of running tests with PyTest.}
\label{f.run_tests}    
\end{figure}

\begin{figure}[!ht]
\centering
\includegraphics[scale=1]{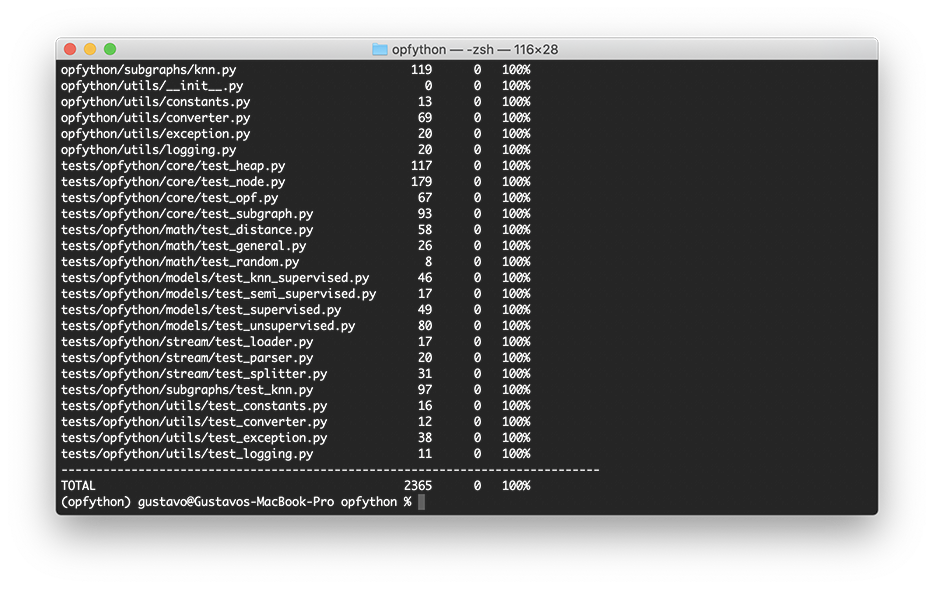}
\caption{Example of code coverage tests with Coverage.}
\label{f.coverage_tests}    
\end{figure}

OPFython is prepared with tests to give a more in-depth analysis of the code. Also, the intention behind any test is to check whether everything is running as demanded or not. Thus, there are two main methods in order to execute the tests:

\begin{itemize}
    \item \textbf{PyTest:} The first method is running the solo command \verb|pytest tests/|, as depicted by Figure~\ref{f.run_tests}. It will fulfill all the implemented tests and return an output indicating whether they succeeded or failed; 

    \item \textbf{Coverage:}, An interesting extension to PyTest is the coverage module. Despite granting the same outputs from PyTest, it will also present a report stating how much the tests cover the code, as illustrated by Figure~\ref{f.coverage_tests}. Its usage is also straightforward: \verb|coverage run -m pytest tests/| and \verb|coverage report -m|. 
\end{itemize}

\section{Applications}
\label{s.applications}

In this section, we explain how to perform a classification task with OPFython, as well as briefly describe the seven pre-loaded applications that are included with the library.

\subsection{Getting Started}
\label{ss.getting_started}

After installing the OPFython library, it is straightforward to use its packages, where seven elementary examples show key features that the library implements. One can refer to the \verb|examples/applications| folder and examine the following files:

\begin{itemize}

\item \textbf{KNN-based supervised OPF training:} \verb|knn_supervised_opf_training.py|;
\item \textbf{Semi-supervised OPF:} \verb|semi_supervised_opf_training.py|;
\item \textbf{Supervised OPF agglomerative learning:} \verb|supervised_opf_agglomerative.py|;
\item \textbf{Supervised OPF learning:} \verb|supervised_opf_learning.py|;
\item \textbf{Supervised OPF with pre-computed distances:} \verb|supervised_opf_pre_computed_distances.py|;
\item \textbf{Supervised OPF pruning:} \verb|supervised_opf_pruning.py|;
\item \textbf{Supervised OPF training:} \verb|supervised_opf_training.py|;
\item \textbf{Unsupervised OPF clustering:} \verb|unsupervised_opf_clustering.py|.

\end{itemize}

\begin{figure}[!ht]
\centering
\includegraphics[scale=1]{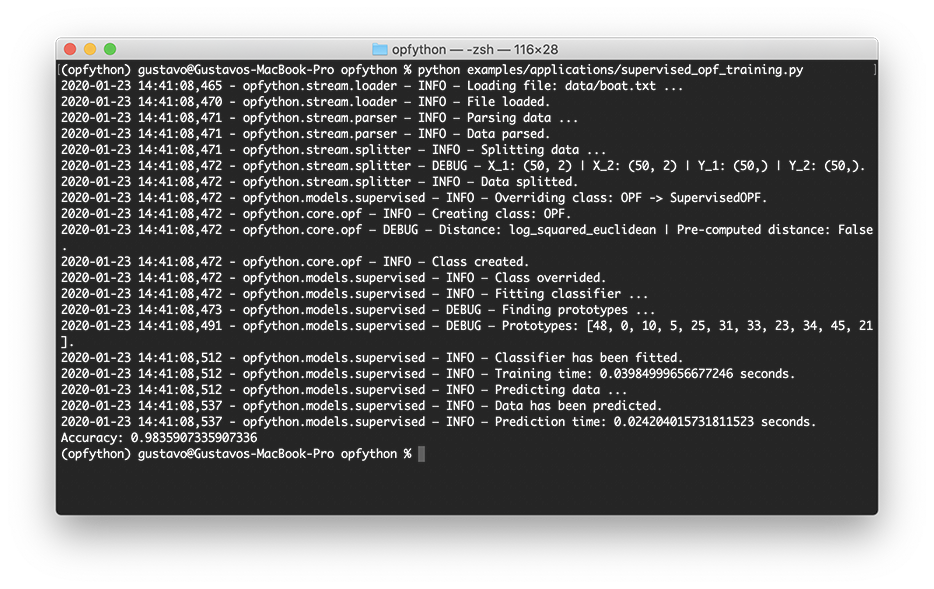}
\caption{Output logs generated by executing an OPFython classification.}
\label{f.get_started}    
\end{figure}

Each example comprises the following pipeline: loading the dataset, parsing the dataset, splitting the dataset, instantiating a classifier, fitting the training data, predicting the validation/test data, and calculating the classifier's performance. Finally, after performing the classification process, it is possible to save the model in a disk-file for further inspection. Figure~\ref{f.get_started} illustrates the output logs generated by an OPFython classification.

The difference between the provided scripts consists of the type of classifier. While supervised classifications attempt to learn a set of classes from particular samples representing them, the unsupervised classification tries to aggregate samples into clusters, i.e., dense regions where samples share some similar traits. As for now, we offer three supervised classifications, e.g., KNN-based supervised OPF, semi-supervised OPF, and supervised OPF, as well as unsupervised classification, the unsupervised OPF. Additionally, we offer three extensions of the supervised OPF, e.g., supervised OPF agglomerative learning (learns from mistakes over the validation set), supervised OPF learning (learns the best classifier over a validation set), and supervised OPF pruning (prunes nodes while maintaining the accuracy).

\subsection{Modeling a New Classification}
\label{ss.function}

In order to model a new classification, some conventional rules need to be comprehended. First of all, the data should be loaded and parsed, which in this case, we will be loading a common dataset known as Boat:

\begin{lstlisting}[language=Python]
import opfython.stream.loader as l     
import opfython.stream.parser as p

# Loading a .txt file to a numpy array     
txt = l.load_txt('data/boat.txt')     
  
# Parsing a pre-loaded numpy array     
X, Y = p.parse_loader(txt)
\end{lstlisting}

Furthermore, if necessary, we can split the data into new sets, such as training and testing, as follows:

\begin{lstlisting}[language=Python]
import opfython.stream.splitter as s    

# Splitting data into training and testing sets     
X_train, X_test, Y_train, Y_test = s.split(X, Y, percentage=0.5, random_state=1)
\end{lstlisting}

Afterward, we can instantiate an OPF classifier:

\begin{lstlisting}[language=Python]
from opfython.models import SupervisedOPF

# Creates a SupervisedOPF instance      
opf = SupervisedOPF(distance='log_squared_euclidean', pre_computed_distance=None)
\end{lstlisting}

Finally, we can fit the classifier and perform new predictions:

\begin{lstlisting}[language=Python]
# Fits training data into the classifier     
opf.fit(X_train, Y_train)

# Predicts new data      
preds = opf.predict(X_test)
\end{lstlisting}

After predicting new samples, it is possible to evaluate the classifier's performance:

\begin{lstlisting}[language=Python]
import opfython.math.general as g    

# Calculating accuracy      
acc = g.opf_accuracy(Y_test, preds)     
\end{lstlisting}
\section{Conclusions}
\label{s.conclusion}

This article introduces an open-source Python-inspired library for handling Optimum-Path Forest classifiers, known as OPFython. Based on an object-oriented paradigm, OPFython provides a modern yet straightforward implementation, allowing users to prototype new OPF-based classifiers swiftly.

The library implements a wide variety of Optimum-Path Forest classifiers, such as supervised, semi-supervised, and unsupervised ones, and auxiliary functions that assist the classifiers' workflow, i.e., distance functions, classification metrics, data processing, errors logging. Additionally, as the original LibOPF thoroughly inspires OPFython's library, it is possible to use the same loading format (OPF file format) and available methods in the original package. Furthermore, OPFython provides a model-saving method, which can be used to pre-train classifiers and retrieve insightful information about the classification procedure.

Regarding future works, we intend to make available more OPF-based classifiers, as well as a visualization package, which will allow users to feed their saved models and furnish charts. Furthermore, we aim to improve our implementations by distributing the calculations, i.e., employing a parallel computing concept, which will hopefully reduce our computational burden.

\section*{Acknowledgments}
The authors are grateful to S\~ao Paulo Research Foundation (FAPESP) grants \#2013/07375-0, \#2014/12236-1, \#2017/25908-6, \#2018/15597-6, and \#2019/02205-5, as well as CNPq grants \#307066/2017-7 and \#427968/2018-6.

\bibliographystyle{unsrt}  
\bibliography{paper}

\end{document}